\documentclass[10pt,letterpaper]{article}
\usepackage[letterpaper,margin=1in]{geometry}
\usepackage[T1]{fontenc}
\usepackage[utf8]{inputenc}
\usepackage{amsmath,amsthm,mathtools}
\usepackage{newtxtext,newtxmath}
\usepackage{graphicx,booktabs,array,tabularx}
\usepackage[table,dvipsnames]{xcolor}
\usepackage{enumitem,microtype}
\usepackage[font=small,labelfont=bf]{caption}
\usepackage{url}
\usepackage[colorlinks=true,citecolor=blue,linkcolor=blue,urlcolor=blue]{hyperref}
\usepackage{bookmark}
\newtheorem{theorem}{Theorem}
\newtheorem{proposition}{Proposition}

\theoremstyle{remark}

\newcommand{\R}{\mathbb{R}}
\newcommand{\T}{\mathbb{T}}
\newcommand{\Rot}{\mathcal{R}}

\newcommand{\softplus}{\operatorname{Softplus}}
\DeclareMathOperator{\diag}{diag}
\DeclareMathOperator{\atanTwo}{atan2}
\setlist{nosep,topsep=4pt}
\graphicspath{{figs/}}
\title{Ananke: Contractive Torus Attractor Networks}
\author{Zhongping Ji}
\date{}
\begin{document}
\maketitle

\begin{abstract}
We introduce Ananke, a representation-learning framework that scaffolds latent representations onto a structured product-torus prior, and its flagship visual backbone realization, Contractive Torus Attractor Networks (CTAN). By factorizing high-dimensional latent spaces into an orthogonal direct sum of two-dimensional phase planes ($\bigoplus_{k=1}^K \R^2$), Ananke coordinates feature updates via a decoupled dual-phase continuous flow: skew-symmetric Hamiltonian transport moves features tangentially along energy level sets to preserve semantic phase invariants, while signed gradient dissipation contracts transverse perturbations normally toward target invariant manifolds. For circular potential families with frozen parameters, logarithmic radial feedback yields the Exact Log-Symplectic Flow (ELSF), an analytical closed-form mapping with exact exponential decay of log-radius error that evaluates in a single forward pass without numerical integration. We establish local input-to-state bounds for level-set deviations and log-radius errors, and characterize the normal hyperbolicity and persistence of the ideal product torus under bounded perturbations. We further formulate the architecture through Lie--Trotter operator splitting, unifying spatial depthwise diffusion with local manifold contraction, and analyze both exact trigonometric flows and hardware-friendly symplectic dual-shear variants. Across natural image benchmarks (CIFAR-100) and clinically challenging endoscopy datasets (Kvasir-v2), CTAN demonstrates exceptional parameter efficiency: an ultra-compact hierarchical model with merely 0.27M parameters achieves 90.52\% accuracy on Kvasir-v2, outperforming 25M+ baselines (ResNet-50, DenseNet-161) by nearly two orders of magnitude in capacity, while scaled variants attain 80.32\% top-1 accuracy on CIFAR-100.
\end{abstract}

\section{Introduction}
Convolutional networks, residual networks, and vision transformers provide effective visual representations by cascading discrete spatial and channel transformations \cite{he2016deep,dosovitskiy2020image,liu2022convnet}. Complementary to static discrete layers, continuous-depth models interpret network depth as physical evolution time governed by ordinary differential equations (ODEs) \cite{chen2018neural}. However, designing continuous dynamical backbones for robust representation learning faces a fundamental dilemma: purely conservative (Hamiltonian or symplectic) flows preserve volume and energy but lack dissipative attraction to filter noise \cite{greydanus2019hamiltonian,zhong2020symplectic}, whereas globally contractive dynamics universally drive all feature trajectories toward a single trivial equilibrium, obliterating inter-class semantic separability.

We resolve this tension by introducing \textbf{Ananke} \footnote{Named after Ananke in Plato’s \emph{Republic}, our framework echoes the Spindle of Necessity through contraction toward structured invariant manifolds.}, a representation-learning framework based on orthogonal symplectic-gradient geometric flows. Rather than presuming that arbitrary data manifolds are globally homeomorphic to a monolithic shape, Ananke adopts a \textbf{Product-Torus Scaffolding Prior}: in analogy to harmonic analysis where orthogonal circular functions serve as a universal basis to decompose signals, we factorize the latent space into an orthogonal direct sum of canonical two-dimensional phase planes ($\bigoplus_{k=1}^K \R^2$, where $D=2K$). Each constituent circle acts as an elementary geometric scaffold, naturally decoupling feature dynamics into transverse gradient dissipation (filtering noise along the radial amplitude) and tangential Hamiltonian transport (encoding semantic variation along the phase). To adapt this canonical base to complex, multimodal distributions, a lightweight spatial context generator dynamically modulates local geometric parameters, effectively tailoring the scaffold to the specific data context.

For isotropic circular potentials, we derive the \textbf{Exact Log-Symplectic Flow (ELSF)}. Under scale-invariant logarithmic feedback, the transverse coordinate $z=\log(r/R)$ linearizes identically into pure exponential decay $\dot z=-\beta z$. This admits an exact, closed-form mapping that evaluates in a single forward pass with zero numerical truncation error, bypassing iterative ODE solvers and step-size stability constraints. Furthermore, we explicitly characterize radial-dependent angular transport (twist dynamics), integrating it in closed form to map transverse deviations into bounded phase offsets.

To scale this formulation into an empirical visual backbone, we introduce \textbf{Contractive Torus Attractor Networks (CTAN)}. CTAN embodies an operator-split realization of continuous reaction--diffusion systems, alternating spatial depthwise convolution (diffusion) with exact closed-form manifold contraction (reaction). 

The principal contributions of this work are:
\begin{enumerate}
\item \textbf{Product-Torus Geometric Scaffolding:} We introduce an orthogonal direct-sum factorization that scaffolds high-dimensional latent spaces into canonical 2D phase planes, establishing a principled, data-adaptive coordinate framework for decoupled radial regulation and phase transport.
\item \textbf{Closed-Form Exact Log-Symplectic Flow (ELSF):} We derive an exact, single-pass closed-form operator for circular geometries under logarithmic feedback, providing an analytical solution that incorporates radial-dependent twist dynamics without numerical overshooting.
\item \textbf{Theoretical Grounding via Contraction and Normal Hyperbolicity:} We establish local input-to-state stability (ISS) bounds for transverse level-set errors and circular log-radii, and prove the normal hyperbolicity and conditional persistence of the uncoupled product torus under bounded perturbations.
\item \textbf{Operator-Split Vision Architecture and Empirical Validation:} We construct hierarchical CTAN backbones integrating spatial diffusion with local geometric reaction across pure and residual topologies. On CIFAR-100 and Kvasir-v2, CTAN demonstrates state-of-the-art parameter efficiency, matching or exceeding heavyweight 25M+ convolutional and transformer baselines with up to $87\times$ fewer parameters.
\end{enumerate}

\section{Related Work}
\label{sec:related_work}
\paragraph{Visual and geometric representations.}
Modern visual backbones predominantly evolve through modular heuristics: deep residual networks stabilize optimization via identity shortcuts \cite{he2016deep}, vision transformers capture long-range token relationships via self-attention \cite{dosovitskiy2020image}, and modernized ConvNets leverage large-kernel depthwise convolutions \cite{liu2022convnet,tolstikhin2021mlp}. Recently, geometric representation learning has explored algebraic structures, such as Clifford algebras and multivector products \cite{ji2026cliffordnet,brandstetter2023clifford,ruhe2023geometric}, and complex-valued neural networks that preserve amplitude--phase relationships \cite{trabelsi2018deep}. The classical manifold hypothesis posits that high-dimensional sensory data concentrate near lower-dimensional submanifolds \cite{fefferman2016testing}. Ananke operationalizes this hypothesis by imposing an explicit product-torus prior, providing structured geometric scaffolding with provable transverse attraction and exact local solvability.

\paragraph{Continuous physical dynamics and operator splitting.}
Interpreting deep architectures as continuous dynamical systems has spurred the development of Neural ODEs \cite{chen2018neural}, Hamiltonian Neural Networks \cite{greydanus2019hamiltonian}, and Symplectic Networks \cite{zhong2020symplectic}. In spatial modeling, PDE-Net connects learned convolutional filters to partial differential operators \cite{long2018pde}. Operator splitting techniques, such as Lie--Trotter and Strang splitting, are classical foundations for separating spatial diffusion from local reaction kinetics in computational physics \cite{strang1968construction}. CTAN bridges these paradigms by structuring vision blocks into an operator-split reaction--diffusion pipeline: spatial depthwise convolution acts as spatial smoothing, while the closed-form ELSF executes exact local manifold contraction.

\paragraph{Contraction theory and invariant manifolds.}
Contraction analysis investigates the convergence between trajectories using matrix logarithmic norms and one-sided Lipschitz conditions \cite{lohmiller1998contraction,bullo2026contraction}. Unlike unconstrained global contraction, which collapses all representations to a single equilibrium, Ananke establishes transverse semicontraction: it enforces exponential contraction along the normal bundle to extinguish off-manifold perturbations, while preserving tangential Hamiltonian transport to sustain semantic discrimination. Furthermore, by framing the product torus within Normally Hyperbolic Invariant Manifold (NHIM) theory \cite{fenichel1971persistence}, we characterize the conditional persistence of the underlying attractor geometry under smooth coupling perturbations.

\section{Methodology}
\label{sec:framework}
\subsection{Paired states and regular level sets}
Let $X\in\R^{B\times D\times H_s\times W_s}$ denote a visual feature tensor with even channel count $D=2K$. Here, the orthogonal direct sum serves as a canonical geometric scaffold: rather than constraining the global data distribution to a rigid topology, each 2D phase plane provides an elementary coordinate frame where amplitude deviations and semantic phase can be regulated independently. Across the spatial lattice, these local fibers are conditioned on the input feature context, adapting the scaffold to the underlying data geometry. At each spatial coordinate $i$, the local latent state vector $x_i\in\R^D$ is factored into an orthogonal direct sum of $K$ independent two-dimensional phase planes:
\begin{equation}
x_i = \bigoplus_{k=1}^K x_{i,k}, \qquad x_{i,k} \in \R^2.
\end{equation}
Within each 2D phase subspace, representations evolve under a conditional reaction flow:
\begin{equation}
\dot x_{i,k}=F_{i,k}(x_{i,k};\eta_{i,k}),\qquad
\mathcal M_\eta=\prod_{i,k}\{x_{i,k}:H_{i,k}(x_{i,k};\eta_{i,k})=0\}.
\label{eq:product_manifold_def}
\end{equation}

\begin{figure}[t]
\centering
\includegraphics[width=\linewidth]{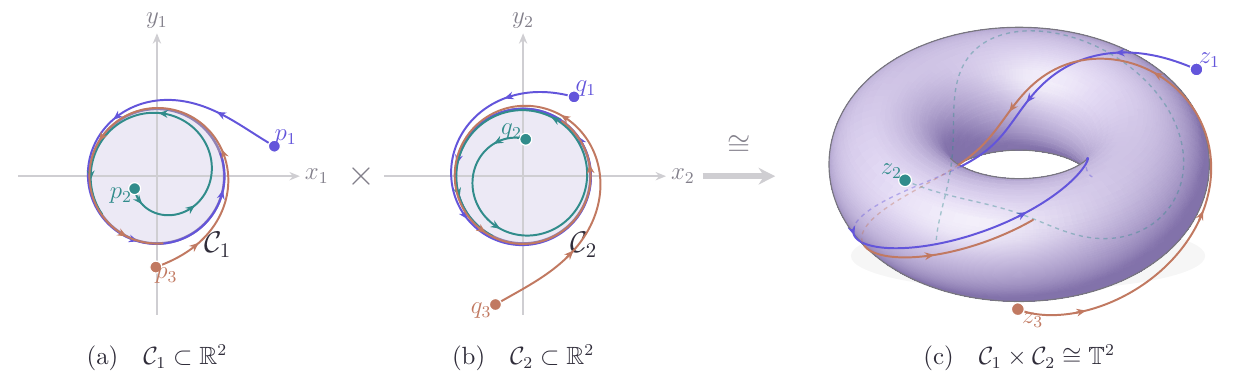}
\caption{Geometric state decomposition and invariant toroidal attractor dynamics in 4D latent space. \textbf{(a, b)} A latent feature vector is orthogonally factored into independent 2D phase planes ($x = x_1 \oplus x_2 \in \R^2 \oplus \R^2$). In each plane, signed gradient dissipation contracts perturbed states ($p_i, q_i$) normally toward an invariant target circle ($\mathcal{C}_1, \mathcal{C}_2$), while skew-symmetric Hamiltonian transport sustains continuous phase rotation. \textbf{(c)} The Cartesian product $\mathcal{C}_1 \times \mathcal{C}_2$ defines an invariant 2-torus $\T^2 \subset \R^4$ (depicted via its canonical embedding in $\R^3$). Composite feature trajectories $z_i = (p_i, q_i)$ are asymptotically contracted from ambient space onto the compact toroidal surface, along which persistent phase orbits encode semantic invariants without representation collapse.}
\label{fig:torus_decomposition}
\end{figure}

The parameter collection $\eta$ is computed once at the block input and held frozen during the reaction step, ensuring decoupled, parallel planar flows without cross-dimensional interference. As visualized in Figure~\ref{fig:torus_decomposition} for a 4D latent subspace ($D=4, K=2$), states in each independent phase plane spiral smoothly toward their respective invariant circular attractors $\mathcal{C}_k \subset \R^2$ under signed dissipation, while their Cartesian product scaffolds representations onto a compact, non-chaotic invariant 2-torus $\T^2 \subset \R^4$.

For an arbitrary constituent pair (omitting spatial and channel indices for brevity), let $H\in C^2(\Omega)$ on an open domain $\Omega\subset\R^2$, and let $g\in C^1(I)$ be a regulatory feedback function satisfying $g(0)=0$ and $u g(u)>0$ for all nonzero $u\in I$. With tangential velocity $\alpha\in\R$ and dissipation rate $\beta>0$, we define the continuous-time symplectic-gradient vector field:
\begin{equation}
F(x)=\underbrace{\alpha J\nabla H(x)}_{\text{Symplectic Tangential Flow}}-\underbrace{\beta g(H(x))\nabla H(x)}_{\text{Signed Gradient Dissipation}},\qquad
J=\begin{pmatrix}0&-1\\1&0\end{pmatrix}.
\label{eq:ananke}
\end{equation}
Here $J$ is the canonical skew-symmetric symplectic matrix satisfying $J^\top=-J=J^{-1}$ and $J^2=-I_2$. When $0$ is a regular value ($\nabla H\ne0$ on $\mathcal M = H^{-1}(0)$), $\mathcal M$ defines a smooth, regular one-dimensional invariant submanifold.

\begin{proposition}[Level-set dissipation and invariance]
\label{prop:dissipation}
Along any trajectory of \eqref{eq:ananke}, the transverse Lyapunov function $V_H(x)=\frac12 H(x)^2$ satisfies:
\begin{equation}
\dot V_H = \langle \nabla V_H(x), F(x) \rangle = -\beta H(x) g(H(x))\|\nabla H(x)\|^2 \le 0.
\label{eq:lyapunov_fun}
\end{equation}
Furthermore, $\dot V_H = 0$ if and only if $x\in\mathcal M$ or $\nabla H(x)=0$. Because $\langle \nabla H, J\nabla H \rangle \equiv 0$, the Hamiltonian component makes identically zero contribution to $\dot V_H$, leaving pure tangential transport along $\mathcal M$ and rendering the regular target level set $\mathcal M$ forward invariant.
\end{proposition}
The proof and sufficient local attraction conditions are detailed in Appendix~\ref{app:energy}.

\subsection{Branch A: exact circular reaction flow}
\label{sec:branch_a_elsf}
When the energy function is parameterized as an isotropic circular potential centered at $c\in\R^2$ with target radius $R>0$:
\begin{equation}
H(x)=\tfrac12(\|x-c\|^2-R^2),\qquad r=\|x-c\|>0,
\qquad g(H)=\log(r/R)=\tfrac12\log\left(1+\tfrac{2H}{R^2}\right),
\label{eq:circ_potential}
\end{equation}
defined on $\Omega=\R^2\setminus\{c\}$. In polar coordinates $x-c=r(\cos\theta,\sin\theta)^\top$, the radial dynamics satisfy the Gompertzian relaxation ODE:
\begin{equation}
\dot r=-\beta r\log(r/R),\qquad \dot z=-\beta z,
\qquad z=\log(r/R).
\label{eq:z_decay}
\end{equation}
Introducing the logarithmic transverse coordinate $z(t)$, the nonlinear radial dynamics linearize identically into pure exponential decay, yielding an exact Lyapunov dissipation rate $\dot V_z = -2\beta V_z$ for $V_z = z^2/2$.

\paragraph{Context modulation and asynchronous twist dynamics.}
To enrich tangential expressiveness, the base angular frequency is modulated by local spatial context, while an optional Arnold twist term couples angular velocity to the transverse radial deviation:
\begin{equation}
\alpha_{\mathrm{eff}}=\alpha_0(1+0.5\tanh p_\alpha),
\qquad \dot\theta=\alpha_{\mathrm{eff}}+a z(t),
\label{eq:twist_ode}
\end{equation}
where $\alpha_0$ is the channel-base frequency, $p_\alpha$ is predicted from neighboring context, and $a=\alpha_{\mathrm{twist}}\in\R$ governs the vortex twist rate.

\begin{theorem}[Exact frozen circular flow operator]
\label{thm:exact}
Fix parameters $c, R, \beta, \alpha_{\mathrm{eff}}, a$ with $R, \beta>0$, and initial state $x_0\ne c$. For any finite horizon $t\ge0$, define:
\begin{align}
q&=e^{-\beta t},&z_0&=\log(\|x_0-c\|/R),\\
r(t)&=R e^{qz_0},&
\vartheta(t)&=\alpha_{\mathrm{eff}}t+\frac a\beta(1-q)z_0.
\label{eq:theta_exact}
\end{align}
The unique analytical solution of the coupled system \eqref{eq:z_decay}--\eqref{eq:twist_ode} is given by:
\begin{equation}
\Phi_t(x_0)=c+\rho(r_0, R, t)\,\Rot_{\vartheta(t)}(x_0-c),
\label{eq:elsf_operator}
\end{equation}
where $\rho = (R/\|x_0-c\|)^{1-q}$ is the radial contraction factor, and $\Rot_\vartheta\in\mathrm{SO}(2)$ is the canonical planar rotation matrix:
\begin{equation}
\Rot_\vartheta = \begin{pmatrix}\cos\vartheta&-\sin\vartheta\\\sin\vartheta&\cos\vartheta\end{pmatrix}.
\end{equation}
The target circle $\|x-c\|=R$ is strictly invariant, with $z(t)=qz_0$. For finite $t$, $\Phi_t$ defines a smooth diffeomorphism of the punctured plane $\R^2\setminus\{c\}$.
\end{theorem}

The analytical integration proof and complex-valued representation are provided in Appendix~\ref{app:exact}. Crucially, \eqref{eq:elsf_operator} evaluates in a single forward pass with zero numerical integration error, eliminating step-size stability constraints.

\paragraph{Exact flow and hardware-friendly algebraic variants}
Theorem~\ref{thm:exact} provides the metrically exact solution $\Rot_\vartheta \in \mathrm{SO}(2)$ along the circular invariant manifold. For resource-constrained edge hardware lacking dedicated Special Function Units (SFUs) for trigonometric transcendentals, one can substitute rational Cayley transforms or area-preserving dual-shear approximations (detailed in Appendix~\ref{app:phase}). Throughout this paper, our primary architecture and all reported benchmarks strictly deploy the exact analytical trigonometric flow $\Rot_\vartheta$.

\subsection{Branch B: regular learned level sets}
\label{sec:branch_b}
Beyond circular geometries, Ananke accommodates expressive learned energy functions integrated via differentiable numerical schemes. To guarantee well-behaved invariant sets, we parameterize $H(x)$ through constrained function classes:
\begin{align}
\text{ICNN Potential:}\quad H_{\rm cvx}(x)&=w^\top\softplus(Ax+b)+\tfrac\lambda2\|x\|^2-h_0,\quad w\ge0,\;\lambda\ge\lambda_{\min}>0,
\label{eq:icnn}\\
\text{SOS Potential:}\quad H_{\rm SOS}(x)&=p(x)^\top M p(x)-R^2,\quad
p(x)=(x_1^2,x_2^2,x_1x_2,x_1,x_2)^\top,\quad
M=W^\top W+\delta I_5,\quad\delta>0.
\label{eq:sos_energy}
\end{align}

\begin{figure}[t]
\centering
\includegraphics[width=\linewidth]{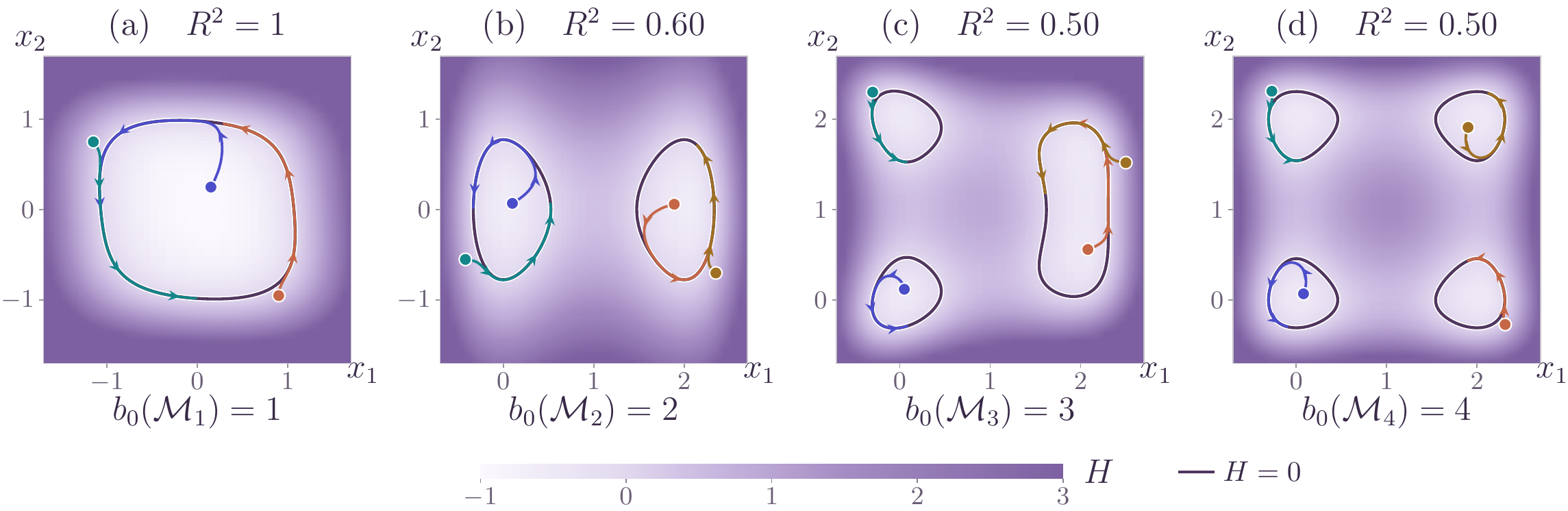}
\caption{Topological transitions and multi-attractor dynamics of the Sum-of-Squares (SOS) algebraic potential. By modulating the Gram matrix and energy offset ($R^2 = 1.0, 0.60, 0.50, 0.50$ across panels (a)--(d)), the zero-level set $\mathcal{M} = \{x : H_{\rm SOS}(x) = 0\}$ (black contours) undergoes Morse-type topological bifurcations, transitioning through configurations indexed by the 0-th Betti number $b_0(\mathcal{M}) \in \{1, 2, 3, 4\}$. Feature trajectories (colored curves) originating from diverse initial states are contracted via transverse gradient dissipation onto locally attracting closed components and transported along tangential Hamiltonian orbits, enabling multimodal clustering within a single 2D phase subspace.}
\label{fig:sos_multicomponent_attractors}
\end{figure}

The strictly positive quadratic terms ($\lambda I$ and $\delta I_5$) enforce radial coercivity ($\lim_{\|x\|\to\infty} H(x) = +\infty$), while choosing $h_0$ strictly above the potential minimum ensures that the zero level set is compact, nonempty, and regular. The quartic SOS algebraic potential exhibits rich Morse-type topological bifurcations: as illustrated in Figure~\ref{fig:sos_multicomponent_attractors}, modulating the algebraic parameters allows the zero-level set to dynamically split into multiple disjoint, locally attracting closed components ($b_0(\mathcal M) \in \{1, 2, 3, 4\}$), naturally realizing multimodal clustering within a single 2D subspace.

To prevent numerical overshooting under discrete integration, we deploy the Algebraic Soft-Clipper (ISRU):
\begin{equation}
g(H) = \frac{\kappa_1 H}{\sqrt{1+(\kappa_2 H)^2}}, \qquad \lim_{H\to\pm\infty} g(H) = \pm \frac{\kappa_1}{\kappa_2},
\end{equation}
which bounds far-field velocity while sustaining non-vanishing polynomial backpropagation gradients. Complete gradient formulations and stability analyses are detailed in Appendix~\ref{app:learned}.

\section{Theoretical Guarantees and Their Scope}
\label{sec:theory}
\subsection{Local energy-error stability}
\begin{theorem}[Local input-to-state bound for level-set error]
\label{thm:energy_iss}
Consider the perturbed flow $\dot x=F(x)+w(t)$ with vector field \eqref{eq:ananke}, and let $U\subset\Omega$ be a compact neighborhood of the regular target manifold $\mathcal M$. Suppose the trajectory remains in $U$ on $[0,T]$ and define:
\begin{equation}
c_U=\inf_{x\in U}\beta\,\frac{g(H(x))}{H(x)}\|\nabla H(x)\|^2>0,
\qquad L_U=\sup_{x\in U}\|\nabla H(x)\|<\infty,
\label{eq:energy_constants}
\end{equation}
where the ratio at $H=0$ is defined continuously as $g'(0)>0$. Then for any essentially bounded perturbation $w$, the transverse error $e(t)=H(x(t))$ satisfies:
\begin{equation}
|H(x(t))|\le e^{-c_Ut}|H(x(0))|
+\frac{L_U}{c_U}(1-e^{-c_Ut})\|w\|_{\infty,[0,t]}, \qquad \forall t\in[0,T].
\label{eq:energy_iss}
\end{equation}
\end{theorem}

Theorem~\ref{thm:energy_iss} establishes an explicit local input-to-state stability (ISS) certificate that mathematically embodies transverse semicontraction: along the normal bundle, transverse perturbations decay exponentially at rate $c_U$ toward an asymptotic error bound proportional to $\|w\|_\infty$, while tangential Hamiltonian transport remains strictly conservative ($\langle \nabla H, J\nabla H \rangle \equiv 0$). This guarantees that off-manifold noise is systematically attenuated without collapsing tangential semantic discrimination. The proof is provided in Appendix~\ref{app:energy}.

\subsection{Sharp input bound for circular log-radius error}
\begin{theorem}[Log-radius error with additive input]
\label{thm:log_iss}
For the frozen circular flow under additive input $w$, suppose the radial trajectory satisfies $r(t)\ge r_{\min}>0$ on $[0,T]$. Then the logarithmic error $z(t)=\log(r(t)/R)$ satisfies:
\begin{equation}
|z(t)|\le e^{-\beta t}|z(0)|+
\frac{1-e^{-\beta t}}{\beta r_{\min}}\|w\|_{\infty,[0,t]},
\qquad 0\le t\le T.
\label{eq:log_iss}
\end{equation}
Furthermore, for fixed bounds $0<r_{\min}<R<r_{\max}$, the compact annulus $\mathcal A = \{x : r_{\min}\le \|x-c\|\le r_{\max}\}$ is strictly forward invariant under the input bound:
\begin{equation}
\|w\|_{\infty}\le
\min\{\beta r_{\min}\log(R/r_{\min}),
\beta r_{\max}\log(r_{\max}/R)\}.
\label{eq:annulus_condition}
\end{equation}
\end{theorem}
Condition~\eqref{eq:annulus_condition} guarantees that the restorative velocity strictly dominates the maximum disturbance at both the inner and outer boundaries, ensuring that representations remain confined within an active annular domain. The proof is detailed in Appendix~\ref{app:log}.

\subsection{Normal hyperbolicity of the ideal product torus}
\begin{theorem}[Conditional persistence of the frozen product torus]
\label{thm:nhim}
Consider $K$ independent circular flows with fixed centers $c_k$, radii $R_k>0$, and contraction rates $\beta_k>0$. Their Cartesian product $\mathcal M_0=\prod_{k=1}^K\{\|x_k-c_k\|=R_k\} \simeq \T^K$ is a compact Normally Hyperbolic Invariant Manifold (NHIM). There exists a fixed tubular neighborhood $V$ of $\mathcal M_0$ and a threshold $\varepsilon_0>0$ such that every $C^2$ vector field $F_\varepsilon$ on $V$ satisfying $\|F_\varepsilon-F_0\|_{C^1(V)}<\varepsilon_0$ admits a nearby invariant manifold $\mathcal M_\varepsilon$ that is $C^1$-diffeomorphic to $\T^K$.
\end{theorem}
Theorem~\ref{thm:nhim} establishes that the unperturbed stable normal contraction rate $\beta_{\min}=\min_k\beta_k>0$ strictly dominates the neutral tangent spectrum. By Fenichel's persistence theorem \cite{fenichel1971persistence}, the underlying toroidal topology persists stably under sufficiently small smooth inter-channel couplings. Appendix~\ref{app:nhim} constructs the invariant splitting in isochron coordinates.

\section{Algorithmic Realization}
\label{sec:algorithmic_realization}
\subsection{Operator-split reaction--diffusion pipeline}
We model visual feature evolution across space and channels via a continuous reaction--diffusion system:
\begin{equation}
\frac{\partial X}{\partial t} = \underbrace{\mathcal D \nabla^2 X}_{\text{Spatial Diffusion}} + \underbrace{F_{\mathrm{local}}(X)}_{\text{Local Manifold Reaction}}.
\label{eq:reaction_diffusion}
\end{equation}
Applying Lie--Trotter operator splitting over time interval $\Delta t$, the continuous PDE decomposes into two sequential stages:
\begin{enumerate}
\item \textbf{Diffusion Step:} Multi-scale spatial depthwise convolution $\mathcal S$ performs local spatial feature smoothing, modeling the heat-diffusion step $\widetilde X = \exp(\Delta t \mathcal D \nabla^2)X_l$.
\item \textbf{Reaction Step:} The single-pass closed-form ELSF operator executes local manifold contraction, advancing features along the target level sets: $X_{l+1} = \Psi_{\Delta t, \eta}(\widetilde X)$.
\end{enumerate}
This operator splitting justifies the elimination of explicit inter-patch ODE coupling terms while preserving the single-pass closed-form solvability of local geometric flows.

\subsection{The context-conditioned torus field}
\label{sec:context}
To allow representations to adapt to multimodal distributions, geometric parameters are predicted dynamically from input features via a lightweight context generator: $\eta = \mathcal G(X) = (c, R, \beta, \alpha_{\mathrm{eff}}, a)$. For a fixed input $X$, parameters are evaluated once and held frozen during the reaction step. Across the spatial grid $V=\{1,\ldots,N\}$, the joint invariant manifold forms an adaptive fiber bundle:
\begin{equation}
\mathcal M_{\mathrm{field}} = \prod_{i=1}^N \mathcal T_i(\eta_i) \simeq \T^{NK}.
\end{equation}
This context conditioning allows the network to dynamically steer local manifold centers and radii to conform to complex semantic boundaries.

\subsection{Multi-scale hierarchy and block variants}
\label{sec:block_topologies}
\begin{figure}[ht]
\centering
\includegraphics[width=\linewidth]{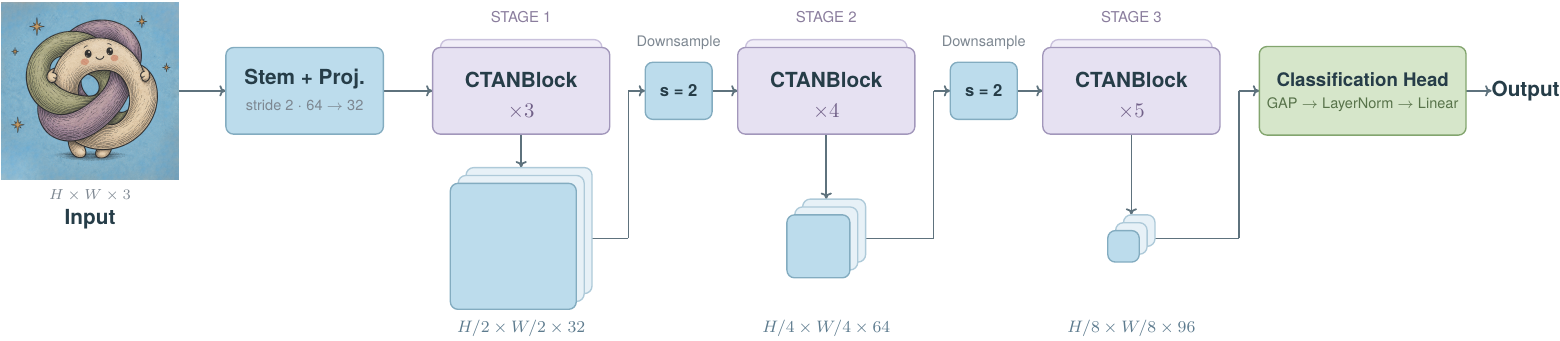}
\caption{The three-stage compact CTAN hierarchy: a stride-two stem, stages of depths $(3,4,5)$ and channel widths $(32,64,96)$, strided transitions, and a pooled classification head. Downsampling changes the state space and does not represent a diffeomorphism between the associated tori.}
\label{fig:hierarchical_architecture}
\end{figure}

As illustrated in Figure~\ref{fig:hierarchical_architecture}, CTAN is structured into a multi-stage pyramidal backbone: an initial strided stem maps RGB inputs into $K_1=D_1/2$ orthogonal phase planes, progressive stages execute intra-stage reaction--diffusion cascades, strided downsampling transitions between scales, and a global pooling head produces class logits.

To cater to distinct computational constraints, we provide two block topologies:
\begin{align}
\text{\textbf{CTAN-Pure:}}\quad x_{\rm sp}&=\mathcal S(\operatorname{Norm}(x_l)),&
x_{l+1}&=\Psi_{t_l,\eta_l}(x_{\rm sp}),
\label{eq:pure}\\
\text{\textbf{CTAN-Res:}}\quad x_{\rm sp}&=\mathcal S(x_l),&
u_l&=W_{\rm in}\operatorname{Norm}(x_{\rm sp}),\notag\\
x_{l+1}&=x_l+\operatorname{DropPath}
\left(\gamma_l\odot W_{\rm out}\Psi_{t_l,\eta_l}(u_l)\right).
\label{eq:res}
\end{align}

In \textsf{CTAN-Res}, the pointwise projections $W_{\rm in}, W_{\rm out} \in \R^{D\times D}$ serve as learnable coordinate adapters. Rather than restricting the phase planes to fixed channel pairs $(2k-1, 2k)$, $W_{\rm in}$ allows the network to learn optimal linear combinations of features as conjugate pairs. Furthermore, while the internal operator $\Psi$ acts on canonical circles, the composition $W_{\rm out} \circ \Psi \circ W_{\rm in}$ flexibly rotates and scales this geometry into oblique ellipsoidal attractors aligned with feature correlations. This gives \textsf{CTAN-Res} greater capacity on complex datasets (Table~\ref{tab:ablation_block_topology}) while preserving the closed-form efficiency of the underlying flow.

\section{Experiments}
\label{sec:experiments}
We evaluate the representation capacity, parameter efficiency, and architectural scaling of CTAN across natural images (CIFAR-100) and clinically demanding endoscopy benchmarks (Kvasir-v2). All models are trained from scratch under standardized optimization protocols. Experimental configurations and dataset details are summarized below.

\subsection{Natural image classification: CIFAR-100}
CIFAR-100 comprises 50,000 training and 10,000 test images across 100 fine-grained categories at $32\times32$ resolution \cite{krizhevsky2009learning}. To rigorously examine the interplay between geometric flows and spatial organization, we instantiate two architectural paradigms across two parameter scales:
\begin{itemize}
    \item  Isotropic Models (CTAN-Iso): A 12-layer uniform backbone maintaining constant spatial resolution and channel width throughout depth:
    \begin{itemize}
        \item CTAN-Iso-1: Width $\text{dim} = 128$ ($K=64$ two-dimensional phase planes), comprising 0.97M parameters;
        \item CTAN-Iso-2: Width $\text{dim} = 160$ ($K=80$ two-dimensional phase planes), comprising 1.49M parameters.
    \end{itemize}
    \item Hierarchical Models (CTAN-Hier): A 3-stage pyramidal hierarchy with patch\_size=1 (preserving spatial resolution in early stages) and progressive downsampling between stages:
    \begin{itemize}
        \item CTAN-Hier-1: stage\_depths=(3, 4, 5), stage\_dims=(96, 128, 160), with shared rotation, yielding 0.91M parameters;
        \item CTAN-Hier-2: stage\_depths=(3, 6, 9), stage\_dims=(96, 144, 192), with shared rotation, scaling to 1.89M parameters.
        \item CTAN-Hier-3: stage\_depths=(3, 6, 9), stage\_dims=(96, 144, 192), with context-conditioned rotation, scaling to 2.13M parameters.
    \end{itemize}
\end{itemize}    
All models are trained from scratch without external pretraining and evaluated against widely adopted lightweight CNNs, deep residual baselines, and state-of-the-art geometric architectures under matched training conditions.

\begin{table}[thbp]
\centering
\caption{Top-1 classification accuracy on CIFAR-100 at $32\times32$ resolution without external pretraining. Best overall result is highlighted in bold; best sub-1M lightweight result is underlined.}
\label{tab:cifar100_results}
\small
\begin{tabular}{@{}lrr@{}}
\toprule
\textbf{Architecture} & \textbf{Parameters} & \textbf{Top-1 accuracy} \\
\midrule
MobileNetV2 \cite{sandler2018mobilenetv2} & 2.3M & 70.90\% \\
ShuffleNetV2 1.5$\times$ \cite{ma2018shufflenet} & 2.6M & 75.95\% \\
ResNet-18 \cite{he2016deep} & 11.2M & 76.75\% \\
ResNet-50 \cite{he2016deep} & 23.7M & 79.14\% \\
CliffordNet-1 \cite{ji2026cliffordnet} & 1.4M & 77.82\% \\
CliffordNet-2 \cite{ji2026cliffordnet} & 2.6M & 79.05\% \\
\textbf{CTAN-Iso-1} (Ours) & 0.97M & 77.65\% \\
\textbf{CTAN-Iso-2} (Ours) & 1.49M & 78.06\% \\
\textbf{CTAN-Hier-1} (Ours) & \underline{\textbf{0.91M}} & \underline{\textbf{78.41\%}} \\
\textbf{CTAN-Hier-2} (Ours) & 1.89M & 79.88\% \\
\textbf{CTAN-Hier-3} (Ours) & 2.13M & \textbf{80.32\%} \\
\bottomrule
\end{tabular}
\end{table}

As reported in Table~\ref{tab:cifar100_results}, CTAN backbones consistently establish superior parameter efficiency:
\begin{itemize}
\item \textbf{Outperforming Heavyweight ResNets with $>11\times$ Compression:} CTAN-Hier-3 attains 80.32\% top-1 accuracy with merely 2.13M parameters, outperforming the massive ResNet-50 baseline (79.14\%, 23.7M parameters) by +1.18\% while utilizing $11.1\times$ fewer parameters. Furthermore, CTAN-Hier-2 (1.89M) surpasses ResNet-50 by +0.74\% with a $12.5\times$ reduction in parameter footprint.
\item \textbf{Hierarchical Multi-Scale Advantage:} Comparing matched parameter scales, CTAN-Hier-1 (0.91M, 78.41\%) surpasses CTAN-Iso-1 (0.97M, 77.65\%) by +0.76\%. Progressively coarse-graining spatial resolution while expanding phase-plane channels ($96\to128\to160$) allows early stages to filter high-frequency noise while reserving deep toroidal dimensions for abstract semantics.
\item \textbf{Dominance over Classic and Geometric Baselines:} CTAN-Hier-1 outperforms classic lightweight CNNs by substantial margins (+7.51\% over MobileNetV2, +2.46\% over ShuffleNetV2 1.5$\times$) and surpasses the geometric algebra model CliffordNet-1 (1.4M, 77.82\%) by +0.59\% with 35\% fewer parameters.
\end{itemize}

\subsection{Gastrointestinal endoscopy: Kvasir-v2}
Kvasir-v2 is a clinically demanding endoscopic dataset containing 8,000 images across eight pathological and anatomical classes \cite{pogorelov2017kvasir,kvasirv2}. Endoscopic imagery presents severe physical artifacts, including intense specular reflections, fluid bubbles, and non-uniform illumination. We evaluate two ultra-compact configurations trained from scratch at $224\times224$ resolution: CTAN-Nano (0.27M parameters) and CTAN-Tiny (1.12M parameters).

\begin{table}[thbp]
\centering
\caption{Diagnostic classification performance on Kvasir-v2 at $224\times224$ resolution trained from scratch. MCC denotes Matthews correlation coefficient.}
\label{tab:kvasir_results}
\small
\begin{tabular}{@{}lrrr@{}}
\toprule
\textbf{Architecture} & \textbf{Parameters} & \textbf{Accuracy} & \textbf{MCC} \\
\midrule
ResNet-50 \cite{he2016deep} & 23.5M & 88.50\% & 0.8689 \\
DenseNet-161 \cite{huang2017dense} & 26.5M & 89.08\% & 0.8756 \\
ViT-Small \cite{dosovitskiy2020image} & 21.7M & 75.50\% & 0.7209 \\
ConvNeXt-Tiny \cite{liu2022convnet} & 27.8M & 77.83\% & 0.7471 \\
EfficientViT \cite{liu2023efficientvit} & 6.58M & 82.83\% & 0.8041 \\
CAN-Nano \cite{ji2026cliffordnet} & 0.36M & 90.33\% & 0.8902 \\
\textbf{CTAN-Nano} & \textbf{0.27M} & \underline{90.52\%} & \underline{0.8943} \\
\textbf{CTAN-Tiny} & 1.12M & \textbf{91.41\%} & \textbf{0.9019} \\
\bottomrule
\end{tabular}
\end{table}

The diagnostic results in Table~\ref{tab:kvasir_results} highlight the strength of contractive geometric priors:
\begin{itemize}
\item \textbf{Extreme Parameter Efficiency ($87\times$ Reduction):} CTAN-Nano attains 90.52\% accuracy and 0.8943 MCC with merely 0.27M parameters, surpassing ResNet-50 (+2.02\% Acc) and DenseNet-161 (+1.44\% Acc) while using less than 1.2\% of their parameters.
\item \textbf{Resistance to Overfitting from Scratch:} Modern high-capacity backbones such as ViT-Small (75.50\%) and ConvNeXt-Tiny (77.83\%) severely degrade when trained from scratch on 8,000 endoscopic images due to overfitting on specular glare. In contrast, CTAN's transverse gradient dissipation systematically contracts off-manifold high-frequency noise into stable tubular neighborhoods, enabling compact capacity to focus exclusively on diagnostic tissue pathology.
\item \textbf{Scalability:} Scaling to CTAN-Tiny (1.12M) yields monotonic gains (+0.89\% Acc, 0.9019 MCC), establishing that the contractive torus formulation scales predictably across channel and depth tiers.
\end{itemize}

\subsection{Block-topology ablation}
Table~\ref{tab:ablation_block_topology} compares the pure flow composition (CTAN-Pure) against the residual-augmented topology (CTAN-Res) under matched stage depths and widths. CTAN-Pure-1 reduces parameter footprint by 48.4\% (0.47M vs.\ 0.91M) while retaining a competitive 75.49\% accuracy on CIFAR-100. Furthermore, enabling context-conditioned dynamic rotation (Pure-2) lifts accuracy to 76.83\% (+1.34\%). This confirms that CTAN-Pure provides an ultra-minimalist inductive filter ideal for resource-constrained edge and medical deployments, whereas CTAN-Res provides the expressive coordinate freedom necessary for scaling to fine-grained visual categories.

\begin{table}[thbp]
\centering
\caption{CIFAR-100 block-topology comparison under fixed stage depths $(3,4,5)$ and dimensions $(96,128,160)$.}
\label{tab:ablation_block_topology}
\small
\begin{tabular}{@{}lccrr@{}}
\toprule
\textbf{Variant} & \textbf{Block Topology} & \textbf{Rotation Mode} & \textbf{Parameters} & \textbf{Accuracy} \\
\midrule
\textsf{CTAN-Pure-1} & Pure Flow Composition & Shared $\alpha_0$ & \textbf{0.47M} & 75.49\% \\
\textsf{CTAN-Pure-2} & Pure Flow Composition & Dynamic $\alpha(\xi)$ & 0.81M & 76.83\% \\
\textsf{CTAN-Res-1}  & Residual-Augmented    & Shared $\alpha_0$ & 0.91M & \textbf{78.41\%} \\
\bottomrule
\end{tabular}
\end{table}

\subsection{Dynamical mechanics and phase realizations}
To isolate the individual contributions of normal dissipation, tangential transport, and discrete phase operators, Table~\ref{tab:ablation_mechanics} presents a systematic ablation on the CTAN-Hier-1 backbone.

\begin{table}[thbp]
\centering
\caption{Mechanics ablation on CTAN-Hier-1 on CIFAR-100. Isolating the contributions of normal logarithmic dissipation, tangential phase transport, and their coupled dynamics.}
\label{tab:ablation_mechanics}
\small
\begin{tabular}{@{}cllllr@{}}
\toprule
\textbf{ID} & \textbf{Update Mode} & \textbf{Radial Dissipation} & \textbf{Tangential Phase} & \textbf{Phase Velocity} & \textbf{Accuracy} \\
\midrule
A & No evolution ($t=0$) & None & None & --- & 73.39\% \\
B & Hard radial projection ($t\to\infty$) & Hard & None & --- & 74.85\% \\
C & Tangential only & None & Trigonometric & Shared $\alpha_0$ & 78.35\% \\
D & Radial only & Soft log-decay & None & --- & 78.75\% \\
E & Tangential only & None & Trigonometric & Dynamic $\alpha(\xi)$ & \underline{79.35\%} \\
F & Full ELSF (Coupled) & Soft log-decay & Trigonometric & Dynamic $\alpha(\xi)$ & \textbf{79.92\%} \\
\bottomrule
\end{tabular}
\end{table}

The empirical findings confirm our theoretical principles across four key dimensions:
\begin{enumerate}
\item \textbf{Necessity of Active Geometric Flows:} Setting $t=0$ (Row A) deactivates the geometric reaction flow, causing performance to plummet to 73.39\% (a severe $-6.67\%$ degradation relative to full flow). This confirms that continuous geometric dynamics constitute the core computational engine rather than cosmetic regularizers.
\item \textbf{Soft Contraction vs.\ Hard Projection:} Enforcing an infinitely rigid projection onto the manifold (Row B, $t\to\infty$, 74.85\%) severely underperforms continuous soft log-decay (Row D, 78.75\%) by $-3.90\%$. This validates our tubular neighborhood analysis in Section~\ref{sec:theory}: hard projection eliminates radial backpropagation gradients ($\partial r_{\rm final}/\partial r_0 \equiv 0$) and discards sample-wise confidence, whereas finite-time soft contraction preserves an active tubular neighborhood $U_\varepsilon$ that maintains smooth gradient flow.
\item \textbf{Context-Adaptive Phase Routing:} Dynamic context-conditioned angular velocity (Row E, 79.35\%) outperforms static shared frequency (Row C, 78.35\%) by $+1.0\%$, demonstrating that spatially heterogeneous phase velocities prevent phase-locking and enhance class margin separation.
\item \textbf{Synergy of Coupled Radial-Tangential Dynamics:} 
Coupling transverse logarithmic dissipation with tangential phase transport (Row F, 79.92\%) achieves the peak performance, surpassing both pure radial contraction (Row D, 78.75\%) and pure tangential rotation (Row E, 79.35\%). This empirical synergy directly validates our core theoretical premise: transverse gradient dissipation purges high-frequency off-manifold noise, while tangential Hamiltonian transport preserves semantic discrimination. Neither component is redundant, and their orthogonal interplay is essential for learning robust, high-density representations.
\end{enumerate}

\section{Discussion and Limitations}
\label{sec:discussion}
The product-torus scaffolding prior provides a structured geometric inductive bias by decomposing high-dimensional representations into orthogonal radial-phase coordinates. This formulation guarantees exact local solvability and provable transverse stability while allowing spatial context networks to adaptively sculpt representation geometries.

\paragraph{Limitations.} While CTAN demonstrates compelling parameter efficiency and mathematical tractability, several operational boundaries remain to be explored:
\begin{enumerate}
\item \textbf{Fixed Pairing Topology:} The current implementation factors channels into fixed coordinate pairs $(x_{2k-1}, x_{2k})$. Exploring data-dependent, dynamic symplectic pairings $J(X)$ presents a promising avenue for multi-channel correlation modeling.
\item \textbf{Evaluation at Scale:} While validated extensively on CIFAR-100 and Kvasir-v2, scaling to web-scale pretraining (e.g., ImageNet-1K) and dense pixel-level prediction (e.g., semantic segmentation) represents an ongoing engineering milestone.
\item \textbf{General Learned Attractors:} Branch B (ICNN and SOS polynomials) accommodates multimodal cluster topologies but requires differentiable numerical solvers, reintroducing step-size considerations during backpropagation.
\end{enumerate}

\section{Conclusion}
In this work, we presented \textbf{Ananke} and its visual realization \textbf{CTAN}, a continuous-time representation learning framework grounded in a product-torus scaffolding prior. By decoupling feature updates into normal gradient dissipation and tangential Hamiltonian transport, CTAN preserves semantic phase invariants while strictly contracting off-manifold disturbances. For circular potentials, the Exact Log-Symplectic Flow (ELSF) provides an exact analytical mapping that evaluates in a single pass with zero truncation error. Framing the backbone through Lie--Trotter operator splitting seamlessly unifies spatial diffusion with local manifold contraction. Empirical evaluations on natural image and clinical endoscopy benchmarks demonstrate that CTAN establishes state-of-the-art parameter efficiency, confirming that structured, context-adaptive torus scaffolding provides a powerful and mathematically transparent foundation for next-generation visual backbones.

\bibliographystyle{unsrt} 
\bibliography{ctan}  

\begin{thebibliography}{10}

\bibitem{he2016deep}
Kaiming He, Xiangyu Zhang, Shaoqing Ren, and Jian Sun.
\newblock Deep residual learning for image recognition.
\newblock In {\em IEEE Conference on Computer Vision and Pattern Recognition (CVPR)}, pages 770--778, 2016.

\bibitem{dosovitskiy2020image}
Alexey Dosovitskiy, Lucas Beyer, Alexander Kolesnikov, Dirk Weissenborn, Xiaohua Zhai, Thomas Unterthiner, Mostafa Dehghani, Matthias Minderer, Georg Heigold, Sylvain Gelly, Jakob Uszkoreit, and Neil Houlsby.
\newblock An image is worth 16x16 words: Transformers for image recognition at scale.
\newblock In {\em International Conference on Learning Representations (ICLR)}, 2021.

\bibitem{liu2022convnet}
Zhuang Liu, Hanzi Mao, Chao-Yuan Wu, Christoph Feichtenhofer, Trevor Darrell, and Saining Xie.
\newblock A convnet for the 2020s.
\newblock In {\em IEEE/CVF Conference on Computer Vision and Pattern Recognition (CVPR)}, pages 11976--11986, 2022.

\bibitem{chen2018neural}
Ricky T.~Q. Chen, Yulia Rubanova, Jesse Bettencourt, and David Duvenaud.
\newblock Neural ordinary differential equations.
\newblock In {\em Advances in Neural Information Processing Systems (NeurIPS)}, volume~31, pages 6571--6583, 2018.

\bibitem{greydanus2019hamiltonian}
Samuel Greydanus, Misko Dzamba, and Jason Yosinski.
\newblock Hamiltonian neural networks.
\newblock In {\em Advances in Neural Information Processing Systems (NeurIPS)}, volume~32, pages 15379--15389, 2019.

\bibitem{zhong2020symplectic}
Yaofeng~Desmond Zhong, Biswadip Dey, and Amit Chakraborty.
\newblock Symplectic ode-net: Learning hamiltonian dynamics with control.
\newblock In {\em International Conference on Learning Representations (ICLR)}, 2020.

\bibitem{tolstikhin2021mlp}
Ilya~O. Tolstikhin, Neil Houlsby, Alexander Kolesnikov, Lucas Beyer, Xiaohua Zhai, Thomas Unterthiner, Jessica Yung, Andreas Steiner, Daniel Keysers, Jakob Uszkoreit, Mario Lucic, and Alexey Dosovitskiy.
\newblock Mlp-mixer: An all-mlp architecture for vision.
\newblock In {\em Advances in Neural Information Processing Systems (NeurIPS)}, volume~34, pages 24261--24272, 2021.

\bibitem{ji2026cliffordnet}
Zhongping Ji.
\newblock Cliffordnet: All you need is geometric algebra.
\newblock {\em arXiv preprint arXiv:2601.06793}, 2026.

\bibitem{brandstetter2023clifford}
Johannes Brandstetter, Rianne van~den Berg, Max Welling, and Jayesh~K. Gupta.
\newblock Clifford neural layers for pde modeling.
\newblock In {\em International Conference on Learning Representations (ICLR)}, 2023.

\bibitem{ruhe2023geometric}
David Ruhe, Jayesh~K. Gupta, Steven de~Keninck, Max Welling, and Johannes Brandstetter.
\newblock Geometric clifford algebra networks.
\newblock In {\em International Conference on Machine Learning (ICML)}, pages 29306--29337, 2023.

\bibitem{trabelsi2018deep}
Chiheb Trabelsi, Olexa Bilaniuk, Ying Zhang, Dmitriy Serdyuk, Sandeep Subramanian, Jo{\~{a}}o~Felipe Santos, Soroush Mehri, Negar Rostamzadeh, Yoshua Bengio, and Christopher~J. Pal.
\newblock Deep complex networks.
\newblock In {\em International Conference on Learning Representations (ICLR)}, 2018.

\bibitem{fefferman2016testing}
Charles Fefferman, Sanjoy Mitter, and Hariharan Narayanan.
\newblock Testing the manifold hypothesis.
\newblock {\em Journal of the American Mathematical Society}, 29(4):983--1049, 2016.

\bibitem{long2018pde}
Zichao Long, Yiping Lu, XianZhong Ma, and Bin Dong.
\newblock Pde-net: Learning pdes from data.
\newblock In {\em International Conference on Machine Learning (ICML)}, pages 3208--3216, 2018.

\bibitem{strang1968construction}
Gilbert Strang.
\newblock On the construction and comparison of difference schemes.
\newblock {\em SIAM Journal on Numerical Analysis}, 5(3):506--517, 1968.

\bibitem{lohmiller1998contraction}
Winfried Lohmiller and Jean-Jacques~E. Slotine.
\newblock On contraction analysis for non-linear systems.
\newblock {\em Automatica}, 34(6):683--696, 1998.

\bibitem{bullo2026contraction}
Francesco Bullo.
\newblock {\em Contraction Theory for Dynamical Systems}.
\newblock Kindle Direct Publishing, 1.3 edition, 2026.

\bibitem{fenichel1971persistence}
Neil Fenichel.
\newblock Persistence and smoothness of invariant manifolds for flows.
\newblock {\em Indiana University Mathematics Journal}, 21(3):193--226, 1971.

\bibitem{krizhevsky2009learning}
Alex Krizhevsky.
\newblock Learning multiple layers of features from tiny images.
\newblock Technical report, University of Toronto, 2009.

\bibitem{sandler2018mobilenetv2}
Mark Sandler, Andrew Howard, Menglong Zhu, Andrey Zhmoginov, and Liang-Chieh Chen.
\newblock Mobilenetv2: Inverted residuals and linear bottlenecks.
\newblock In {\em Proceedings of the IEEE Conference on Computer Vision and Pattern Recognition (CVPR)}, pages 4510--4520, 2018.

\bibitem{ma2018shufflenet}
Ningning Ma, Xiangyu Zhang, Hai-Tao Zheng, and Jian Sun.
\newblock Shufflenet v2: Practical guidelines for efficient cnn architecture design.
\newblock In {\em Proceedings of the European Conference on Computer Vision (ECCV)}, pages 116--131, 2018.

\bibitem{pogorelov2017kvasir}
Konstantin Pogorelov, Kristin~Ranheim Randel, Carsten Griwodz, Sigrun~Losada Eskeland, Thomas de~Lange, Dag Johansen, Concetto Spampinato, Duc-Tien Dang-Nguyen, Mathias Lux, Peter~Thelin Schmidt, Michael Riegler, and P{\aa}l Halvorsen.
\newblock Kvasir: A multi-class image dataset for computer aided gastrointestinal disease detection.
\newblock In {\em Proceedings of the 8th ACM on Multimedia Systems Conference}, pages 164--169, 2017.

\bibitem{kvasirv2}
{Simula Research Laboratory}.
\newblock Kvasir dataset, version 2.
\newblock Dataset resource, 2017.
\newblock \url{https://datasets.simula.no/kvasir/}.

\bibitem{huang2017dense}
Gao Huang, Zhuang Liu, Laurens van~der Maaten, and Kilian~Q. Weinberger.
\newblock Densely connected convolutional networks.
\newblock In {\em Proceedings of the IEEE Conference on Computer Vision and Pattern Recognition (CVPR)}, pages 4700--4708, 2017.

\bibitem{liu2023efficientvit}
Xinyu Liu, Houwen Peng, Ningxin Zheng, Yuqing Yang, Han Hu, and Yixuan Yuan.
\newblock Efficientvit: Memory efficient vision transformer with cascaded group attention.
\newblock In {\em Proceedings of the IEEE/CVF Conference on Computer Vision and Pattern Recognition (CVPR)}, 2023.

\bibitem{amos2017icnn}
Brandon Amos, Lei Xu, and J.~Zico Kolter.
\newblock Input convex neural networks.
\newblock In {\em Proceedings of the 34th International Conference on Machine Learning}, volume~70 of {\em Proceedings of Machine Learning Research}, pages 146--155. PMLR, 2017.

\bibitem{duncan2017nonreversible}
Andrew~B. Duncan, Grigorios~A. Pavliotis, and Konstantinos~C. Zygalakis.
\newblock Nonreversible langevin samplers: Splitting schemes, analysis and implementation.
\newblock {\em arXiv preprint arXiv:1701.04247}, 2017.

\end{thebibliography}

\clearpage
\appendix
\numberwithin{equation}{section}
\section{Level-Set Dissipation and Local Error Bounds}
\label{app:energy}
\subsection{Proof of Proposition~\ref{prop:dissipation}}
Since $J^\top=-J$, $v^\top Jv=0$ for every real vector $v$. Therefore
\begin{equation}
\frac{d}{dt}H(x(t))
=\nabla H^\top(\alpha J\nabla H-\beta g(H)\nabla H)
=-\beta g(H)\|\nabla H\|^2.
\label{eq:energy_derivative}
\end{equation}
Multiplication by $H$ gives \eqref{eq:lyapunov_fun}. Its equality set is the union of $H^{-1}(0)$ and the critical set of $H$. On a regular zero level the vector field is tangent, and uniqueness of the flow gives invariance for the duration of its existence. When the regular zero level is compact, the restricted smooth vector field is complete. 

\subsection{Proof of Theorem~\ref{thm:energy_iss}}
Let $e(t)=H(x(t))$. With input,
\begin{equation}
\dot e=-\beta g(e)\|\nabla H\|^2+\nabla H^\top w.
\end{equation}
At $e\ne0$, multiplication by $\operatorname{sign}(e)$, followed by the definitions of $c_U,L_U$, gives
\begin{equation}
\frac{d}{dt}|e|\le-c_U|e|+L_U\|w(t)\|.
\end{equation}
The corresponding upper derivative inequality also holds at zero (or almost everywhere for measurable inputs). Integrating the scalar comparison equation gives the stronger convolution form
\begin{equation}
|e(t)|\le e^{-c_Ut}|e(0)|+
L_U\int_0^t e^{-c_U(t-\tau)}\|w(\tau)\|\,d\tau,
\end{equation}
from which \eqref{eq:energy_iss} follows. 

\subsection{When positive local constants exist}
Suppose $\mathcal M$ is compact and regular and $g'(0)>0$. By continuity, there exists a sufficiently small compact tubular neighborhood $U$ on which $\|\nabla H\|\ge m>0$ and $g(H)/H\ge\mu>0$, ensuring $c_U\ge\beta\mu m^2 > 0$. An upper gradient bound $L_U < \infty$ exists by compactness. Furthermore, on this tubular neighborhood, the energy error $|H(x)|$ is comparable up to fixed positive constants to the geometric distance $\operatorname{dist}(x,\mathcal M)$, thereby translating the energy ISS bound into a direct distance-to-set stability certificate.

\section{Exact Circular Flow, Phase Coordinates, and Derivatives}
\label{app:exact}
\subsection{Proof of Theorem~\ref{thm:exact}}
For $r>0$, division of the radial equation by $r$ gives $\dot z=-\beta z$, hence $z(t)=e^{-\beta t}z_0$ and $r(t)=Re^{qz_0}$. Integrating the angular equation gives
\begin{equation}
\theta(t)-\theta_0=\alpha_{\mathrm{eff}}t+
a z_0\int_0^t e^{-\beta\tau}\,d\tau
=\alpha_{\mathrm{eff}}t+\frac a\beta(1-q)z_0.
\end{equation}
Reconstruction in Cartesian coordinates yields \eqref{eq:elsf_operator}. The radius remains strictly positive for every finite time. The $(z,\theta)$ update, with angle understood modulo $2\pi$, is invertible:
\begin{equation}
z_0=z(t)/q,\qquad
\theta_0=\theta(t)-\alpha_{\mathrm{eff}}t-\frac a\beta(1-q)z_0.
\end{equation}
It is smooth on the punctured plane. At $z_0=0$ the radius remains $R$, proving invariance. The same formulas apply componentwise to the fixed, decoupled product flow.

\subsection{Phase dynamics and isochron coordinates}
For two trajectories governed by \eqref{eq:z_decay}--\eqref{eq:twist_ode} with shared parameters, the angular deviation evolves as $\Delta\theta(t) = \Delta\theta_0 + \frac{a}{\beta}(1-q)\Delta z_0$. To decouple the twist-induced radial coupling, we define the isochron phase coordinate:
\begin{equation}
\psi = \theta + \frac{a}{\beta} z \pmod{2\pi}.
\label{eq:isochron}
\end{equation}
Differentiating yields $\dot\psi = \dot\theta + \frac{a}{\beta}\dot z = (\alpha_{\mathrm{eff}} + az) - az = \alpha_{\mathrm{eff}}$. Consequently, the coupled vortex dynamics linearize into a rigid translation in $\psi$-coordinates, preserving isochron phase differences $\Delta\psi(t) \equiv \Delta\psi_0$ across time.

\subsection{Complex notation and local product topology}
Using a complex relative state $u=(x^{(1)}-c^{(1)})+\mathrm{i}(x^{(2)}-c^{(2)})$, distinct from the real log-error $z$, the exact ODE is
\begin{equation}
\dot u=\left[-\beta\log(|u|/R)
+\mathrm{i}\{\alpha_{\mathrm{eff}}+a\log(|u|/R)\}\right]u.
\end{equation}
Fixed-parameter reaction is equivariant under a constant rotation of $u$. This is a $U(1)$ action and does not establish invariance of the classifier to phase changes or local gauge equivariance of spatial convolutions.

For fixed positive radii, $\iota(\theta)_k=c_k+R_k(\cos\theta_k,\sin\theta_k)$ smoothly embeds $\T^K$ in $\R^{2K}$. On $U_\delta=\{\max_k|\log(r_k/R_k)|<\delta\}$, the deformation
\begin{equation}
(r_k,\theta_k)\longmapsto
(R_k\exp[(1-s)\log(r_k/R_k)],\theta_k),\qquad 0\le s\le1,
\end{equation}
retracts $U_\delta$ onto the product torus. This is a topological fact about a fixed neighborhood, not a claim that arbitrary feature distributions or network layers preserve its fundamental group. For equal radii $R_k=R$ and zero centers, the torus lies on the sphere of radius $R\sqrt K$; its codimension within that sphere is $K-1$, so it is a hypersurface only when $K=2$.

\section{Proof of the Circular Input Bound}
\label{app:log}
Additive input changes the radial and log-radius equations to
\begin{equation}
\dot r=-\beta r\log(r/R)+e_r^\top w,
\qquad \dot z=-\beta z+\frac{e_r^\top w}{r}.
\end{equation}
If $r\ge r_{\min}$, integration yields
\begin{equation}
z(t)=e^{-\beta t}z(0)+\int_0^t e^{-\beta(t-\tau)}
\frac{e_r(\tau)^\top w(\tau)}{r(\tau)}\,d\tau.
\end{equation}
Taking absolute values proves \eqref{eq:log_iss}. At the inner boundary $r=r_{\min}$, the unperturbed outward radial speed is $\beta r_{\min}\log(R/r_{\min})$; at the outer boundary the inward speed has magnitude $\beta r_{\max}\log(r_{\max}/R)$. Condition~\eqref{eq:annulus_condition} prevents an input from pushing either boundary outward from the closed annulus, proving forward invariance. Smoothness on this compact annulus and bounded measurable input give continued existence there. 

\section{Invariant Splitting and Conditional Persistence}
\label{app:nhim}
\subsection{Proof of Theorem~\ref{thm:nhim}}
For the fixed product flow, let $z_k=\log(r_k/R_k)$ and $\psi_k=\theta_k+(a_k/\beta_k)z_k$. In these coordinates,
\begin{equation}
\dot z_k=-\beta_k z_k,\qquad
\dot\psi_k=\alpha_{\mathrm{eff},k}.
\end{equation}
The torus is $z=0$. The tangent bundle splits into the tangent directions $\delta z=0$ and an invariant stable complement $\delta\psi=0$. In $(z_k,\theta_k)$ coordinates the latter is spanned by $(1,-a_k/\beta_k)$, rather than by the geometric radial direction when $a_k\ne0$. Under the time-$t$ flow, stable variations contract by $e^{-\beta_k t}$ while tangent variations are unchanged in the adapted norm. This gives normal domination with rate $\beta_{\min}>0$.

The torus is compact and lies away from all centers. The ideal vector field is smooth on a neighborhood of it. Persistence of normally hyperbolic invariant manifolds \cite{fenichel1971persistence} therefore gives a nearby normally attracting invariant manifold under sufficiently small $C^1$ perturbations of the vector field, for the $C^2$ perturbed vector fields specified in Theorem~\ref{thm:nhim}. All norms and smallness conditions are taken on a fixed neighborhood of the ideal torus. 

Theorem~\ref{thm:nhim} characterizes the structural stability of the underlying continuous reaction dynamics. In practical network implementations, this continuous invariant manifold serves as a robust geometric inductive prior, while surrounding discrete operations (e.g., spatial mixing and residual projections) provide the necessary representation capacity for downstream visual tasks.

\section{Hardware-Friendly Algebraic Phase Approximations}
\label{app:phase}

While the canonical trigonometric operator $\Rot_\vartheta$ in Theorem~\ref{thm:exact} provides the metrically exact solution, evaluating transcendental functions ($\cos, \sin$) utilizes GPU Special Function Units (SFUs). For resource-constrained edge hardware lacking dedicated SFU pipelines, this appendix characterizes two algebraic alternatives: the rational Cayley transform and the symplectic dual-shear operator.

\subsection{Rational Cayley transform}
The Cayley map parameterizes planar rotations via Weierstrass half-angle rational arithmetic ($u = \vartheta/2$):
\begin{equation}
C(\vartheta) = \frac{1}{1+u^2} \begin{pmatrix} 1-u^2 & -2u \\ 2u & 1-u^2 \end{pmatrix}.
\label{eq:cayley}
\end{equation}
Because $(1-u^2)^2 + (2u)^2 \equiv (1+u^2)^2$, $C(\vartheta)$ strictly preserves the Euclidean norm ($\|C(\vartheta)x\| \equiv \|x\|$) in exact arithmetic, bypassing all SFU calls.

The trigonometric half-angle identity reveals that $C(\vartheta) = \Rot_{\phi(\vartheta)}$ with effective rotation angle $\phi(\vartheta) = 2\arctan(\vartheta/2)$. Taylor expansion yields:
\begin{equation}
\phi(\vartheta) = 2\arctan(\vartheta/2) = \vartheta - \frac{\vartheta^3}{12} + \mathcal O(\vartheta^5).
\end{equation}
Thus, Cayley rotation induces a cubic phase distortion $\Delta\theta = \mathcal O(\vartheta^3)$ relative to the linear physical horizon. In neural networks where angular velocities are learned, this acts as a smooth alternative parameterization of $\mathrm{SO}(2)$ rather than a strict numerical integration error.

\subsection{Symplectic dual-shear operator}
The dual-shear operator $M(s)$ is formulated as the composition of two parabolic nilpotent shears ($s = \vartheta$):
\begin{equation}
M(s) = \begin{pmatrix} 1 & -s \\ 0 & 1 \end{pmatrix} \begin{pmatrix} 1 & 0 \\ s & 1 \end{pmatrix} = \begin{pmatrix} 1-s^2 & -s \\ s & 1 \end{pmatrix}.
\label{eq:shear_matrix}
\end{equation}
In hardware execution, $M(s)x$ evaluates via two in-place Fused Multiply-Add (FMA) steps without transcendental operations:
\begin{equation}
y_{\rm mid} = y_0 + s\cdot x_0, \qquad x' = x_0 - s\cdot y_{\rm mid}, \qquad y' = y_{\rm mid}.
\end{equation}

\paragraph{Area conservation and elliptic invariants.}
Because $\det M(s) = (1-s^2)(1) - (-s)(s) \equiv 1$, dual-shear is strictly area-preserving ($M(s) \in \mathrm{Sp}(2, \R)$). Defining the metric matrix $Q_s = \left(\begin{smallmatrix} 1 & s/2 \\ s/2 & 1 \end{smallmatrix}\right)$, direct multiplication confirms:
\begin{equation}
M(s)^\top Q_s M(s) = Q_s.
\end{equation}
The eigenvalues of $Q_s$ are $1 \pm s/2$, so $Q_s$ is positive-definite for $|s| < 2$. Thus, $M(s)$ preserves the positive-definite quadratic form:
\begin{equation}
\widehat H_s(u_1, u_2) = \tfrac{1}{2} u^\top Q_s u = \tfrac{1}{2}(u_1^2 + u_2^2) + \tfrac{s}{2} u_1 u_2,
\label{eq:shear_energy}
\end{equation}
which defines concentric ellipses tilted at $45^\circ$. For $|s| < 2$, $\operatorname{tr}(M(s)) = 2-s^2 \in (-2, 2)$, ensuring that its eigenvalues lie strictly on the unit circle ($|\lambda| \equiv 1$).

\paragraph{Infinitesimal generator.}
For $0 < |s| < 2$, setting $\phi = \arccos(1 - s^2/2) \in (0, \pi)$, the matrix admits the real representation $M(s) = \cos\phi\,I + s J Q_s$. Its principal matrix logarithm is:
\begin{equation}
\log M(s) = \frac{\phi s}{\sin\phi} J Q_s = s J - \frac{s^2}{2} \diag(1, -1) + \mathcal O(s^3).
\label{eq:shear_log}
\end{equation}
This reveals that $M(s)$ is the finite flow of a canonical rotation $s J$ perturbed by a second-order hyperbolic squeeze $-\frac{s^2}{2} \diag(1, -1)$. While dual-shear induces a slight radial oscillation relative to a perfect circle, its pure FMA execution makes it an attractive algebraic surrogate for latency-critical deployments.

\section{Learned Potentials and Numerical Integration}
\label{app:learned}
\subsection{A convex potential with a regular compact level set}
Let $f(x)=w^\top\softplus(Ax+b)+\lambda\|x\|^2/2$ with the constraints of \eqref{eq:icnn}. If $\sigma$ denotes the logistic function and $v=Ax+b$, then
\begin{align}
\nabla f(x)&=A^\top(w\odot\sigma(v))+\lambda x,\\
\nabla^2 f(x)&=A^\top\diag(w\odot\sigma(v)\odot(1-\sigma(v)))A+\lambda I
\succeq\lambda_{\min}I.
\end{align}
Hence $f$ is strongly convex and coercive, with a unique minimizer. If $h_0>\min f$, the level $f=h_0$ is nonempty, compact, and contains no critical point. These are sufficient conditions for the regularity assumed in the local theory. The input-convex construction follows the parameter-constraint principle of ICNNs \cite{amos2017icnn}. 

\subsection{Derivatives of the SOS family}
For \eqref{eq:sos_energy},
\begin{equation}
p(x)^\top M p(x)\ge\delta\|p(x)\|^2
\ge\delta(x_1^4+x_2^4)\ge\tfrac\delta2\|x\|^4,
\end{equation}
so the potential is coercive. With zero-based components $q_j$ of $q=Mp(x)$, the gradient is
\begin{equation}
\nabla H_{\rm SOS}(x)=2\begin{pmatrix}
2q_0x_1+q_2x_2+q_3\\2q_1x_2+q_2x_1+q_4
\end{pmatrix}.
\end{equation}

\subsection{Feedback saturation and solver stability}
For the algebraic feedback,
\begin{equation}
g(H)=\frac{\kappa_1H}{\sqrt{1+(\kappa_2H)^2}},\quad
g'(H)=\frac{\kappa_1}{[1+(\kappa_2H)^2]^{3/2}},\quad
\lim_{H\to\pm\infty}g(H)=\pm\frac{\kappa_1}{\kappa_2}.
\end{equation}
Its derivative decays algebraically and still tends to zero. The slope and saturation amplitude depend jointly on these coefficients; an explicitly reparameterized form $g(H)=aH/\sqrt{1+(aH/b)^2}$ has slope $a$ and saturation magnitude $b$. Neither form bounds the complete vector field because
\begin{equation}
\|F(x)\|=\sqrt{\alpha^2+\beta^2g(H(x))^2}\,\|\nabla H(x)\|.
\end{equation}

\section{Theoretical Extensions and Future Directions}
\label{app:extensions}
\subsection{Intrinsic phase distances and Riemannian metric heads}
For a fixed product torus with radii $R_k$, the induced flat Riemannian metric gives:
\begin{equation}
d(\theta,\theta_c)=\left[\sum_{k=1}^K R_k^2
\atanTwo\!\left(\sin(\theta_k-\theta_{c,k}),
\cos(\theta_k-\theta_{c,k})\right)^2\right]^{1/2},
\end{equation}
providing a natural geodesic loss for classifying semantic phases on the quotient group $\T^K \simeq \R^K/2\pi\mathbb Z^K$.

\subsection{Multimodal phase locking via Kuramoto dynamics}
An auxiliary modality (e.g., text prompt) can supply conditioning context. For a phase model $\dot\theta_{\rm vis}=\alpha_{\rm vis}+\kappa\sin(\theta_{\rm text}-\theta_{\rm vis})$ with $\dot\theta_{\rm text}=\alpha_{\rm text}$ and coupling $\kappa>0$, the phase difference $\delta=\theta_{\rm vis}-\theta_{\rm text}$ obeys $\dot\delta=\Delta\alpha-\kappa\sin\delta$. Stable frequency locking is certified when $|\Delta\alpha|<\kappa$, providing a physical foundation for cross-modal alignment.

\subsection{Complex and K\"ahler geometry connections}
Identifying $\R^{2K}\simeq\mathbb C^K$ with canonical complex structure $J\leftrightarrow \mathrm{i}$ allows unifying conservative transport and dissipative contraction via Wirtinger derivatives of a single real-valued potential $K(z,\bar z)$:
\begin{equation}
\frac{dz_k}{dt} = \mathrm{i}\alpha \frac{\partial K}{\partial \bar z_k} - \beta g(H) \frac{\partial K}{\partial \bar z_k}.
\end{equation}
Exploring dynamically generated period matrices into the Siegel modular variety presents a rich mathematical frontier for complex manifold representation learning.

\subsection{Stochastic extensions and generative Langevin dynamics}
Augmenting the deterministic flow with Brownian diffusion $\sigma\,dB_t$ transforms the system into an It\^o SDE: $dX=F(X)\,dt+\sigma\,dB_t$. Defining the potential $U(x)=\beta\int_0^{H(x)}g(s)\,ds$ and divergence-free drift $b(x)=\alpha J\nabla H(x)$ yields $\nabla\cdot b=0$ and $b\cdot\nabla U=0$. The stationary Fokker--Planck equation admits the exact Gibbs invariant density:
\begin{equation}
p_*(x)=Z^{-1}e^{-2U(x)/\sigma^2}.
\end{equation}
This establishes a theoretical bridge connecting contractive torus attractors with nonreversible Langevin samplers \cite{duncan2017nonreversible} and continuous-time generative flow matching.

\end{document}